\documentclass[runningheads, envcountsame, a4paper]{llncs}  
\usepackage{llncsdoc}
\usepackage{wrapfig}
\usepackage{amsmath,amssymb}
\usepackage{graphicx}
\usepackage[ruled,linesnumbered]{algorithm2e}
\usepackage{algorithmic}
\usepackage{paralist} 
\usepackage{color}
\usepackage{multirow}
\usepackage{rotating}
\usepackage[mathscr]{eucal} 
\usepackage{amsbsy} 
\usepackage{subfigure}
\usepackage{booktabs}
\usepackage{xcolor}
\usepackage{tikz}
\usetikzlibrary{snakes}
\usepackage{multirow}
\usepackage{epstopdf}
\usepackage{balance}
\usepackage{tablefootnote}
\usepackage{empheq} 
\usepackage{moresize}
\usepackage{enumitem}
\setlist{nolistsep}
\usepackage{sidecap}
\usepackage{mdframed}
\newcommand{\tensor}[1]{\underline{ \mathbf{#1} }}
\usepackage[hyphens]{url}
\usepackage{microtype}

\newcounter{ALC@tempcntr}
\newcommand{\hide}[1]{}

\newcommand{\ben}{\begin{enumerate*}}
\newcommand{\een}{\end{enumerate*}}
\newcommand{\bit}{\begin{itemize*}}
\newcommand{\eit}{\end{itemize*}}

\titlerunning{Concept Drift in Streaming Tensor Decomposition}
\authorrunning{R. Pasricha\and E. Gujral, and E.E. Papalexakis}

\newcommand{\method}{\textit{SeekAndDestroy}\xspace}
\newcommand{\conceptoverlap}{\text{Concept Overlap}\xspace}
\newcommand{\newconcept}{\text{New Concept}\xspace}
\newcommand{\missingconcept}{\text{Missing Concept}\xspace}
\newcommand{\runningrank}{\text{Running Rank}\xspace}
\newcommand{\rr}{\text{runningRank}\xspace}
\newcommand{\conOverlap}{\text{conceptOverlap}\xspace}
\newcommand{\newcon}{\text{newConcept}\xspace}

\begin{document}
\author{Ravdeep Pasricha\and Ekta Gujral\and Evangelos E. Papalexakis}
\institute{Department of Computer Science and Engineering\\
					University of California Riverside\\
				900 University Avenue, Riverside, CA, USA \\
					\email{$\{$rpasr001, egujr001$\}$@ucr.edu, and epapalex@cs.ucr.edu}}
\title{Identifying and Alleviating Concept Drift in Streaming Tensor Decomposition}

\maketitle

\setcounter{footnote}{0} 
\begin{abstract}
Tensor decompositions are used in various data mining applications from social network to medical applications and are extremely useful in discovering latent structures or {\em concepts} in the data.
Many real-world applications are dynamic in nature and so are their data. To deal with this dynamic nature of data,  there exist a variety of online tensor decomposition algorithms. A central assumption in all those algorithms is that the number of latent concepts remains fixed throughout the entire stream. However, this need not be the case. Every incoming batch in the stream may have a different number of latent concepts, and the difference in latent concepts from one tensor batch to another can provide insights into how our findings in a particular application behave and deviate over time. 
In this paper, we define ``concept'' and ``concept drift'' in the context of streaming tensor decomposition, as the manifestation of the variability of latent concepts throughout the stream. Furthermore, we  introduce \method\footnote{The method name is after (and a tribute to) Metallica's song from their first album (who also owns the copyright for the name)}, an algorithm  that detects concept drift in streaming tensor decomposition and is able to produce results robust to that drift. To the best of our knowledge, this is the first work that investigates concept drift in streaming tensor decomposition. 
We extensively evaluate \method on synthetic datasets, which exhibit a wide variety of realistic drift. Our experiments demonstrate the effectiveness of \method, both in the detection of concept drift and in the alleviation of its effects, producing results with similar quality to decomposing the entire tensor in one shot.
Additionally, in real datasets, \method outperforms other streaming baselines, while discovering novel useful components.
\end{abstract}

\section{Introduction}
\label{sec:intro}
Data comes in many shapes and sizes. Many real world applications deal with data that is multi-aspect (or multi-dimensional) in nature. An example of multi-aspect data would be interactions between different users in a social network over period of time. Interactions like who messages whom, who liked whose posts or who shared (re-tweet) whose post. This can be modeled as a three-mode tensor, user-user being two modes of the tensor and time being the third mode, where each data point can be considered as an interaction between two users. 

Tensor decomposition has been used in many data mining applications and is an extremely useful tool for finding latent structures in tensor in an unsupervised fashion. There exist a wide  variety of tensor decomposition models and algorithms available, interested readers can refer to \cite{doi:10.1137/07070111X,papalexakis2017tensors} for details.
In this paper, our main focus is on CP/PARAFAC decomposition \cite{PARAFAC} (henceforth refered to as CP for brevity), which decomposes a tensor into a sum of rank-one tensors, each one being a latent factor (or {\em concept}) in the data. CP has been widely used in many applications, due to its ability to uniquely uncover latent components in a variety of unsupervised multi-aspect data mining applications \cite{papalexakis2017tensors}.

In today's world data is not static, data keeps on evolving over time. In real world applications like stock market and e-commerce websites hundred of transaction (if not thousands) takes place every second, or in applications like social media where every second, thousands of new interactions take place forming new communities of users who interact with each other. In this example, we consider each {\em community} of people within the graph as a {\em concept}.

There has been a considerable amount of work in dealing with online or streaming CP decomposition \cite{zhou2016accelerating,gujral2017sambaten,nion2009adaptive}, where the goal is to absorb the updates to the tensor in the already computed decomposition, as they arrive, and avoid recomputing the decomposition every time new data arrives. However, despite the already existing work in the literature, a central issue has been left, to the best of our knowledge, entirely unexplored. All of the existing online/streaming tensor decomposition literature assumes that the concepts in the data (whose number is equal to the rank of the decomposition) remains {\em fixed} throughout the lifetime of the application. What happens if the number of components changes? What if a new component is introduced, or an existing component splits into two or more new components? This is an instance of {\em concept drift} in unsupervised tensor analysis, and this paper is a look at this problem from first principles.

Our contributions in this paper are the following:
\begin{itemize}[noitemsep]
	\item {\bf Characterizing concept drift in streaming tensors}: We define concept and concept drift in time evolving tensors and provide a quantitative method to measure the concept drift.
	\item {\bf Algorithm for detecting and alleviating concept drift in streaming tensor decomposition}: We provide an algorithm which detects drift in the streaming data and also updates the previous decomposition without any assumption on the rank of the tensor. 
	\item {\bf Experimental evaluation on real \& synthetic data}: We extensively evaluate our method on both synthetic and real datasets and out-perform state of the art methods in cases where the rank is not known a priori  and perform on par in other cases.
	\item {\bf Reproducibility:} Our implementation is made publicly available\footnote{\url{https://github.com/ravdeep003/conceptDrift}} for reproducibility of experiments.
\end{itemize}

\hide{The rest of the paper is structured as follows. In section 2, we outline the background on streaming tensor decomposition and problem definition. We present our proposed method in section 3. Section 4 describes evaluation of our proposed method on synthetic and real world datasets. In section 5 we discuss what little work that have been done previously on this problem and finally we conclude the paper in section 6.}

\section{Problem Formulation}
\label{sec:problem}

\subsection{Tensor Definition and Notations}
\textbf{Tensor} $\tensor{X}$ is collection of stacked matrices ($\mathbf{X}_1,\mathbf{X}_2,\dots \mathbf{X}_K$) with dimension   $\mathbb{R}^{I \times J\times K}$, where $I$ and $J$ represents rows and columns of matrix and $K$ represents number of views. In other words, a tensor is a higher order abstraction of a matrix. For simplicity, we call the term ``dimension'' as ``mode'' of tensor, where ``modes'' are the numbers of views used to index the tensor. The rank($\tensor{X}$) is the minimum number of rank-1 tensors computed from its latent components which are required to re-produce $\tensor{X}$ as their sum.  Table \ref{table:t2} represents the notations used throughout the paper. 
\begin{table}[h!]
	\ssmall
	\begin{center}
		\begin{tabular}{ |c|c| }
			\hline
			Symbols & Definition \\ 
			\hline
			\hline
			
			$\tensor{X},\mathbf{X}, \mathbf{x},x$ & Tensor, Matrix, Column vector, Scalar \\ 
			\hline
			$\mathbb{R}$ & Set of Real Numbers  \\ 
			\hline
			$\circ$ & Outer product  \\ 
			\hline
			$\lVert \mathbf{A} \rVert_F, \| \mathbf{a} \|_2$& Frobenius norm, $\ell_2$ norm \\
		\hide{	\hline 
			$\mathbf{x}(I)$ & Spanning the elements of $\mathbf{x}$ in indices $\in I$ \\ 
			\hline
			$\mathbf{x}(:)$ & Spanning all elements of $\mathbf{x}$\\ }
			\hline
			$\mathbf{X}(:,r)$ &$ r^{th}$ column of $\mathbf{X}$  \\ 
		\hide{	\hline
			$\mathbf{X}(r,:)$ & $ r^{th}$ row of $\mathbf{X}$  \\ 
			\hline
			$\otimes $& Kronecker product\\}
			\hline
			$\odot$ & Khatri-Rao product (column-wise Kronecker product \cite{papalexakis2017tensors})\\
			\hline
		\end{tabular}
		\bigskip
		\caption{{Table of symbols and their description}}
		\label{table:t2}
	\end{center}
	\vspace{-0.2in}
\end{table}

\textbf{Tensor Batch:} A batch is a (N-1)-mode partition of tensor $\tensor{X}  \in \mathbb{R}^{I \times J\times K}$ where size is varied only in one mode and other modes remain unchanged. Here, tensor $\tensor{X}_{new}$ is of dimension $\mathbb{R}^{I \times J\times t_{new}}$ and existing tensor $\tensor{X}_{old}$ is of dimension $\mathbb{R}^{I \times J\times t_{old}}$. The full tensor $\tensor{X} =[\tensor{X}_{old} ;  \tensor{X}_{new}]$ where its temporal mode $K=t_{old} +t_{new}$. The tensor $\tensor{X}$ can be partitioned into horizontal $\tensor{X}$(I,:,:) , lateral $\tensor{X}$(:,J,:), and frontal $\tensor{X}$(:,:,K) mode.

\textbf{CP decomposition:} The most popular and extensively used tensor decompositions is the Canonical Polyadic or CANDECOMP/PARAFAC decomposition, referred to as CP decomposition henceforth. Given a 3-mode tensor $\tensor{X}$ of dimension $\mathbb{R}^{I \times J\times K}$, and  rank at most $R$
can be written  $$\tensor{X}=\sum_{r=1}^{R} (a_r \odot b_r \odot c_r) \iff \tensor{X}(i,j,k) = \sum_{r=1}^{R}A(i,r)B(j,r)C(k,r)$$ $ \forall$  $i \in \{1,2, \dots,I\}$, $j\in \{1,2, \dots,J\}$ , $k \in \{1,2, \dots,K\}$ and   $\mathbf{A} \in \mathbb{R}^{I \times R} , \mathbf{B} \in \mathbb{R}^{J \times R}$ and $\mathbf{C} \in \mathbb{R}^{K \times R}$. For tensor approximation, we adopted minimizing least square criteria as $\mathcal{L}  \approx \min\frac{1}{2}||\tensor{X}  - \mathbf{A}(\mathbf{C} \odot \mathbf{B})^T||_F^2$ where $||\tensor{X}||_F^2$ is the sum of squares of its all elements and $||.||_F$ is \text{\em {Frobenius} } (norm). The CP model is nonconvex in $\mathbf{A}, \mathbf{B}$ and $\mathbf{C}$. We refer interested readers  to popular surveys \cite{doi:10.1137/07070111X,papalexakis2017tensors} on tensor decompositions and its applications for more details.

\subsection{Problem Definition}

Let us consider a social media network like Facebook, where a large number of users ($\approx 684K$) update information every single minute, and Twitter, where about $\approx 100K$ users tweet every minute\footnote{\url{https://mashable.com/2012/06/22/data-created-every-minute/}}. Here, we have interactions arriving continuously at high velocity, where each interaction consists of User Id, Tag Ids , Device, and Location information etc. How can we capture such dynamic user interactions? How to identify concepts which can signify a potential newly emerging community,  complete disappearance of interactions, or a merging of one or more communities to a single one? When using tensors to represent such dynamically evolving data, our problem falls under ``streaming'' or ``online'' tensor analysis. Decomposing streaming or online tensors is challenging task, and concept drift in incoming data makes the problem significantly more difficult, especially in applications where we care about characterizing the concepts in the data, in addition to merely approximating the streaming tensor adequately.

Before we conceptualize the problem that our paper deals with, we define certain terms which are necessary to set up the problem. Consider $\tensor{X}$ and $\tensor{Y}$ be two incremental batches of a streaming tensors of rank $R$ and $F$ respectively. Let $\tensor{X}$ be the initial tensor at time $t_0$ and $\tensor{Y}$ be the batch of the streaming tensor which arrives at time $t_1$ such as $t_1 > t_0$. The CP decomposition for these two tensors is given as follows:
\begin{equation}
\tensor{X} \thickapprox \sum_{r=1}^R \mathbf{A}(:,r) \circ \mathbf{B}(:,r)\circ \mathbf{C}(:,r)
\end{equation}

\begin{equation}
\tensor{Y} \thickapprox \sum_{r=1}^F \mathbf{A}(:,r) \circ \mathbf{B}(:,r)\circ \mathbf{C}(:,r)
\end{equation}	
\noindent{\textbf{Concept}:} In case of tensors, we define {\em concept} as one latent component; a sum of $R$ such components make up the tensor. In above equations tensor $\tensor{X}$ and $\tensor{Y}$ has \textit{R} and \textit{F} concepts respectively.

\begin{figure}[!ht]
	\vspace{-0.1in}
	\begin{center}
		\includegraphics[clip,width = 0.55\textwidth]{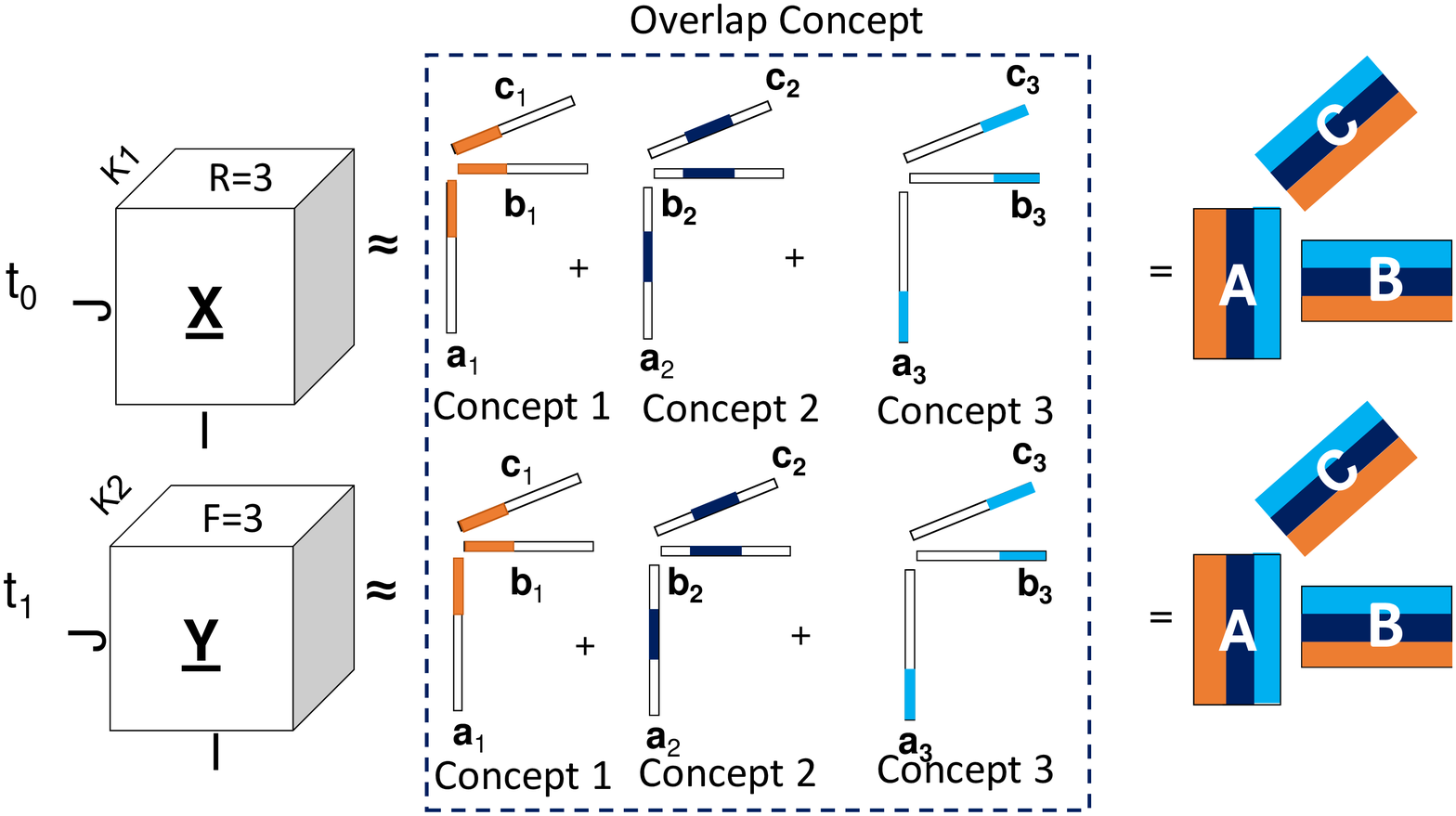}
		\caption{{Complete overlap of concepts}}
		\label{fig:overlap_concept}
	\end{center}
	\vspace{-0.2in}
\end{figure}

\noindent{\textbf{\conceptoverlap}:} We define {\em concept overlap} as the set of latent concepts that are common or shared between two streaming CP decompositions.  Consider Figure \ref{fig:overlap_concept} where $R$ and $F$ both are equal to three, which means both tensors $\tensor{X}$ and $\tensor{Y}$ have three concepts. Each concept of $\tensor{X}$ corresponds to each concept of $\tensor{Y}$. This means that there are three concepts that overlap between $\tensor{X}$ and $\tensor{Y}$. The minimum and maximum number of concept overlaps between two tensors can be zero and $\min(R,F)$ respectively. Thus, the value of concept overlap lies between 0 and $\min(R,F)$.
In Section \ref{sec:method} we propose an algorithm for detecting such overlap.

\begin{equation}
0 \le \conceptoverlap \le \min(R,F)
\end{equation}

\begin{figure}[!ht]
	\vspace{-0.1in}
	\begin{center}
		\includegraphics[clip,width = 0.45\textwidth]{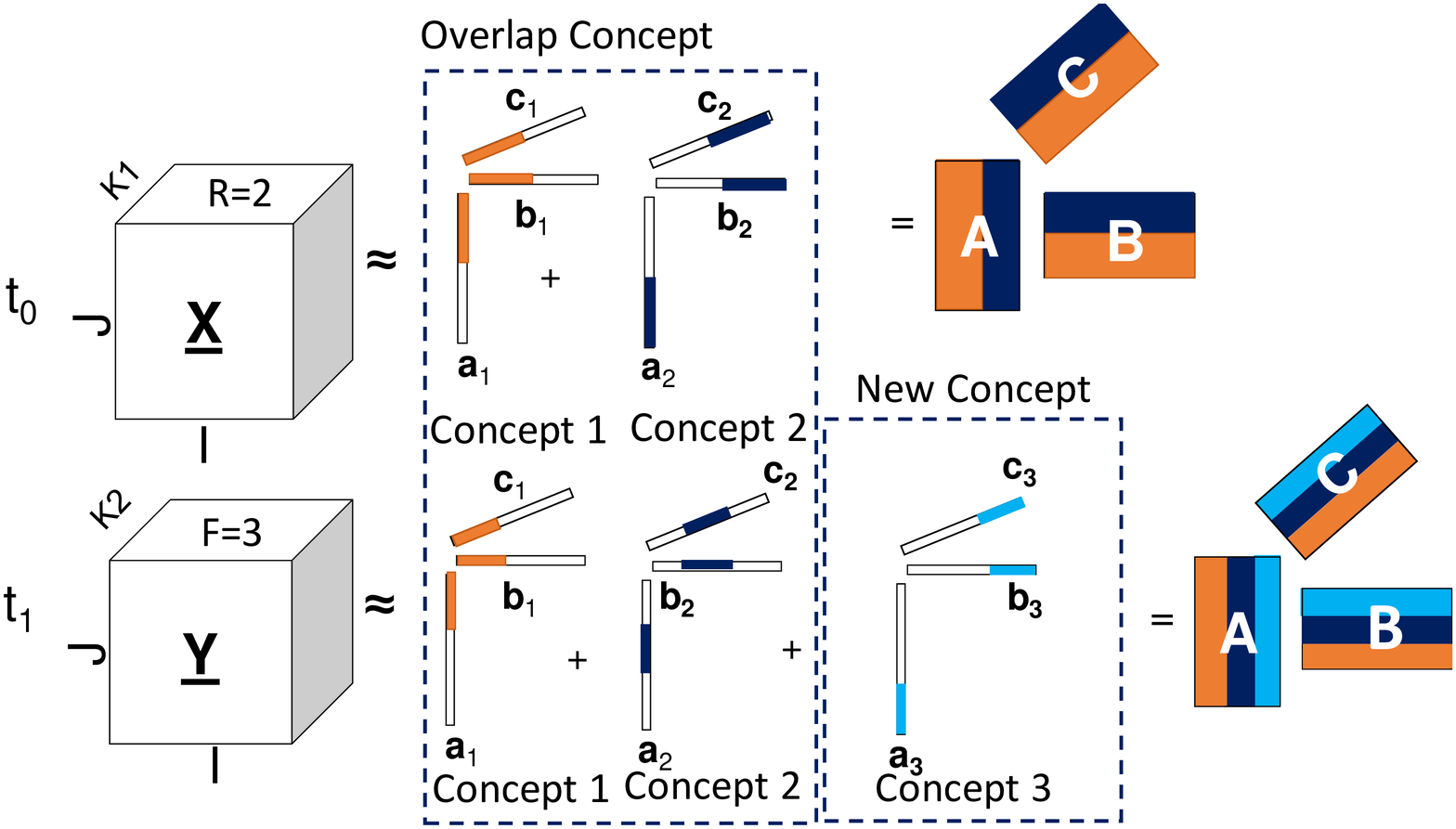}
			\includegraphics[clip,width = 0.45\textwidth]{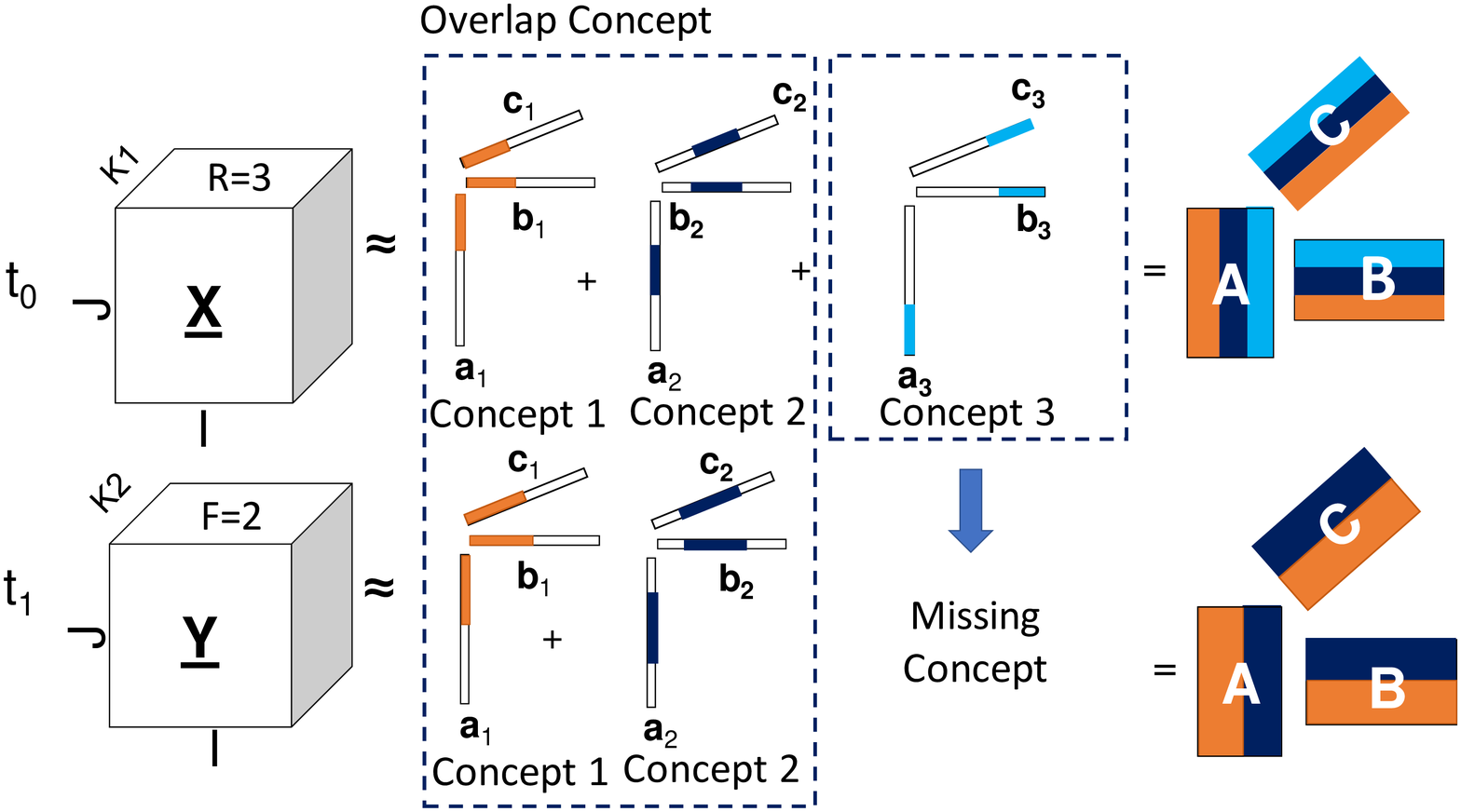}
		\caption{{(a) Concept Appears \ \ \    (b) Concept disappears}}
		\label{fig:new_concept}
	\end{center}
	\vspace{-0.2in}
\end{figure}

\noindent{\textbf{\newconcept}:} If there exists a set of concepts which are not similar to any of the concepts already present in the most recent tensor batch, we call all such concepts in that set as {\em new concepts}.  Consider Figure \ref{fig:new_concept}$(a)$,  where $\tensor{X}$ has two concepts $(R=2)$ and $\tensor{Y}$ has three concepts $(F=3)$. We see that at time $t_1$ tensor $\tensor{Y}$ batch has three concepts, out of which, two match with tensor $\tensor{X}$ concepts and one concept(namely concept 3) does not match with any concept of $\tensor{X}$. In this scenario we say that concept $1$ and $2$ are {\em overlapping concepts} and concept $3$ is a {\em new concept}.

\noindent{\textbf{\missingconcept}:} If there exists a set of concepts which was present at time $t_0$, but was missing at future time $t_1$, we call the concepts in the set {\em missing concepts}. For example, consider Figure \ref{fig:new_concept}$(b)$, at time $t_0$, the CP decomposition of $\tensor{X}$ has three concepts, and at time $t_1$ CP decomposition of $\tensor{Y}$ has two concepts. Two concepts of $\tensor{X}$ and $\tensor{Y}$ match with each other and one concept, present at $t_0$, is missing at $t_1$; we label that concept, as {\em missing concept}.   \\
\hide{
 \begin{figure}[!ht]
 	\vspace{-0.1in}
 	\begin{center}
 		\includegraphics[clip,trim=0.1cm 3.0cm 0.5cm 3.0cm,width = 0.85\textwidth]{FIG/missingConcept.pdf}
 		\caption{{Concept disappears}}
 		\label{fig:missing_concept}
 	\end{center}
 	\vspace{-0.2in}
 \end{figure}
}
\textbf{\runningrank}: \runningrank(\rr) at time $t$ is defined as the total number of unique concepts (or latent components) seen until time $t$. Running Rank is different from tensor rank of a tensor batch. It may or may not be equal to rank of the current tensor batch. Consider Figure \ref{fig:overlap_concept}, \rr at time $t_1$ is three, since the total unique number of concepts seen until $t_1$ is three. Similarly \rr of Figure \ref{fig:new_concept}$(b)$ at time $t_1$ is three, even though rank of $\tensor{Y}$ is two, since the number unique concepts seen until $t_1$ is three.

Let us assume rank of the initial tensor batch $\tensor{X}$ at time $t_0$ is $R$ and rank of the subsequent tensor batch $\tensor{Y}$ at time $t_1$ is $F$. Then \rr at time $t_1$ is sum of running rank at ${t_0}$ and number of new concepts discovered from ${t_0}$ to ${t_1}$. At time ${t_0}$ running rank is equal to initial rank of the tensor batch in this case $R$.

\begin{equation}
\rr_{t_1} =	\rr_{t_0}+ num(\newcon)_{t_1-t_0}
\end{equation}

\noindent{\textbf{Concept Drift}:} Concept drift is usually defined in terms of supervised learning \cite{bifet2011handling,Webb2016,DBLP:journals/corr/WebbLPG17}. In \cite{Webb2016}, authors define concept drift in unsupervised learning as the change in probability distribution of a random variable over time. We define concept drift in the context of latent concepts, which is based on rank of the tensor batch. We first give an intuitive description of concept in terms of running rank, and then define concept drift.

\noindent{\textbf{Intuition}:} Consider running rank at time $t_1$ be $\rr_{t_1}$ and running at time $t_2$ be $\rr_{t_2}$. If $\rr_{t_1}$ is not equal to $\rr_{t_2}$, then there is a concept drift i.e. either a new concept has appeared, or a concept has disappeared. However, this definition does not capture every single case. Assume if $\rr_{t_1}$ is equal to $\rr_{t_2}$. In this case, there is no drift only when there is a complete overlap. However there may be concept drift present even if $\rr_{t_1}$ is equal to $\rr_{t_2}$, since a concept might disappear while \rr remains the same.

\noindent{\textbf{Definition}:} Whenever a new concept appears, a concept disappears, or both from time $t1$ to $t2$, this phenomenon is defined as {\em concept drift}.

In a streaming tensor application, a tensor batch arrives at regular intervals of time. Before we decompose a tensor batch to get latent concepts, we need to know the rank of the tensor. Finding tensor rank is a hard problem \cite{haastad1990tensor} and it is beyond the scope of this paper. There has been considerable amount of work which approximates rank of a tensor\cite{papalexakis2016automatic,morup2009automatic}. In this paper we employ AutoTen \cite{papalexakis2016automatic} to compute a low rank of a tensor. As new advances in tensor rank estimation happen, our proposed method will also benefit.

\begin{figure}[!ht]
	\vspace{-0.1in}
	\begin{center}
		\includegraphics[clip,width = 0.95\textwidth]{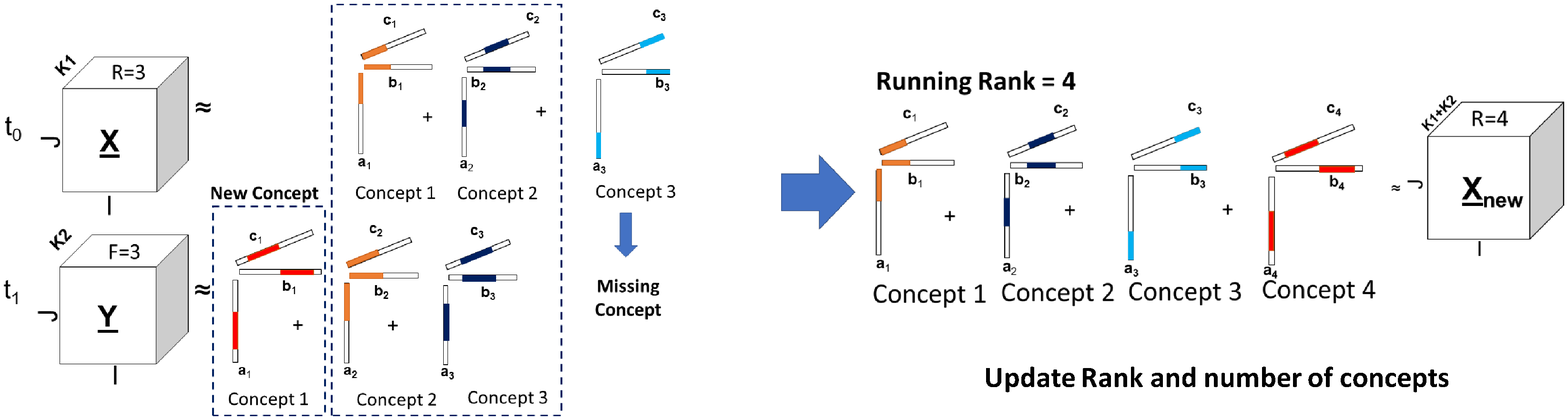}
		\caption{{Problem formulation}}
		\label{fig:complete_problem}
	\end{center}
	\vspace{-0.2in}
\end{figure}

\begin{mdframed}[linecolor=red!60!black,backgroundcolor=gray!20,linewidth=2pt,topline=false,rightline=false, leftline=false] 
	\begin{problem}
{\bf Given} (a) tensor $\tensor{X}$ of dimensions $I \times J \times K_1$ and rank $R$, (b) $\tensor{Y}$ of dimensions $I \times J \times K_2$ of rank $F$ at time $t_0$ and $t_1$ respectively as shown in figure \ref{fig:complete_problem}. Compute $\tensor{X}_{new}$ of dimension $I\times J\times (K_1+K_2)$ of rank equal to \rr at time $t_1$ as shown in equation $(5)$ using factor matrices of $\tensor{X} $ and $ \tensor{Y}$.
\begin{equation}
\tensor{X}_{{new}_{t_1}} \thickapprox \sum_{r=1}^\rr \mathbf{A}(:,r) \circ \mathbf{B}(:,r)\circ \mathbf{C}(:,r)
\end{equation}
\end{problem}
\end{mdframed}

\section{Proposed Method}
\label{sec:method}

 Consider a social media application where thousands of connections are formed every second, for example, who follows whom or who interacts with whom. These connections formed can be viewed as forming communities. Over a period of time communities disappear, new communities appear or some communities re-appear after sometime. Number of communities at any given point of time is dynamic. There is no way of knowing what communities will appear or disappear in future. When this data stream is captured as a tensor, communities refer to latent concepts and appearing and disappearing of communities over a period of a time is referred to as concept drift. Here we need a dynamic way of figuring out number of communities in a tensor batch rather than assuming constant number of communities in all tensor batches.

To the best of our knowledge, there is no algorithmic approach that detects concept drift in streaming tensor decomposition. As we mentioned in Section \ref{sec:intro}, there has been considerable amount of work \cite{gujral2017sambaten,zhou2016accelerating,nion2009adaptive} which deals with streaming tensor data and applies batch decomposition on incoming slices and combine the results. But these methods don't take change of rank in consideration, which could reveal new latent concept in the data sets. Even if we know the rank(latent concept) of the complete tensor, the tensor batches of that tensor might not have same rank as the complete tensor.

In this paper we propose \method, a streaming CP decomposition algorithm that does not assume rank is fixed. \method detects the rank of every incoming batch in order to decompose it, and finally, updates the existing decomposition after detecting and alleviating concept drift, as defined in Section \ref{sec:problem}.

An integral part of \method is detecting different concepts and identifying concept drift in streaming tensor. In order to do this successfully, we need to solve following problems:
\begin{itemize}
	\item[\textbf{P1:}] Finding the rank of a tensor batch.
	\item[\textbf{P2:}] Finding \newconcept, \conceptoverlap and \missingconcept between two consecutive tensor batch decomposition.
	\item[\textbf{P3:}] Updating the factor matrices to incorporate the new and missing concepts along with concept overlaps.
\end{itemize}


\textbf{Finding Number of Latent Concepts}: Finding the rank of the tensor is beyond the scope of this paper, thus we employ AutoTen \cite{papalexakis2016automatic}. Furthermore, in Section \ref{sec:experiments}, we perform our experiments on synthetic data where we know the rank  (and use that information as given to us by an ``oracle'') and repeat those experiments using AutoTen, comparing the error between them; the gap in quality signifies room for improvement that \method will reap, if rank estimation is solved more accurately in the future.

\textbf{Finding Concept Overlap}: Given a rank of tensor batch, we compute its latent components using CP decomposition. 
Consider Figure \ref{fig:complete_problem} as an example. At time ${t_1}$, the number of latent concepts we computed is represented by $F$, and we already had $R$ components before new batch $\tensor{Y}$ arrived. In this scenario, there could be three possible cases: (1) $R = F$ (2) $R > F$ (3) $R < F$.

For each one of the cases mentioned above, there may be new concepts appear at $t_1$, or concepts disappear from $t_0$ to $t_1$, or there could be shared concepts between two decompositions. In Figure \ref{fig:complete_problem}. we see that, even though $R$ is equal to $F$, we have one new concept, one missing concept and two shared/overlapping concepts. Now, at time $t_1$, we have four unique concepts, which means our \rr at ${t_1}$ is four.

\begin{center}
	\begin{algorithm} [!ht]
		\caption{\method for Detecting \& Alleviating Concept Drift}
		\label{Algorithm: Seek And Destroy}
		\begin{algorithmic} [1]
		\REQUIRE Tensor $\tensor{X}_{new}$ of size $I \times J \times K_{new}$, Factor matrices $\mathbf{A}_{old},\mathbf{B}_{old}, \mathbf{C}_{old}$ of size $I \times R$, $J \times R$ and $K_{old} \times R$ respectively, runningRank, mode.
		\ENSURE Factor matrices $\mathbf{A}_{updated}, \mathbf{B}_{updated}, \mathbf{C}_{updated}$ of size $I \times \rr$, $J \times \rr$ and $(K_{new}+K_{old}) \times \rr$, $\boldsymbol{\rho}$, $ \rr$.

		\STATE $batchRank \leftarrow getRankAutoten(\tensor{X}_{new}, runningRank)$
		\STATE $[\mathbf{A}, \mathbf{B}, \mathbf{C},\boldsymbol{\lambda}] = $ CP$\left( \tensor{X}_{new}, batchRank \right)$.
		\STATE $\mathbf{colA}, \mathbf{colB}, \mathbf{colC} \leftarrow$ Compute Column Normalization of $\mathbf{A}, \mathbf{B}, \mathbf{C}$.
		\STATE $\mathbf{normMatA}, \mathbf{normMatB}, \mathbf{normMatC} \leftarrow$ Absorb $\boldsymbol{\lambda}$ and Normalize $\mathbf{A}, \mathbf{B}, \mathbf{C}$.
		\STATE $rhoVal \leftarrow colA\ .*\ colB\ .*\ colC$
		\STATE $[\newcon, \conOverlap, overlapConceptOld] \leftarrow findConceptOverlap(\mathbf{A}_{old},\mathbf{normMatA})$
	    \IF{\newcon} 
	   \STATE  $\rr \leftarrow \rr + len(\newcon)$
	    \STATE $\mathbf{Aupdated} \leftarrow \begin{bmatrix}
	    \mathbf{A}_{old} \ \mathbf{normMatA}(:,\newcon)\\
	    \end{bmatrix}$ 
	    
	    \STATE  $\mathbf{Bupdated} \leftarrow \begin{bmatrix}
	    \mathbf{B}_{old} \ \mathbf{normMatB}(:,\newcon)\\
	    \end{bmatrix}$ 
	    
	    \STATE  $\mathbf{Cupdated} \leftarrow$  update $\mathbf{C}$  depending on the \newconcept, \\ \conceptoverlap, overlapConceptOld indices and $\rr$
	    \ELSE
	    \STATE $\mathbf{Aupdated} \leftarrow \mathbf{A}_{old}$
	    \STATE  $\mathbf{Bupdated} \leftarrow \mathbf{B}_{old}$ 
	    \STATE  $\mathbf{Cupdated} \leftarrow$  update $\mathbf{C}$ depending on the \conceptoverlap, overlapConceptOld indices and $\rr$
	    \ENDIF
	    \STATE Update $\rho$ depending on the \newconcept and \conceptoverlap indices
	    \IF{\newcon or $(len(\newcon) + len(\conOverlap) < \rr)$}
	    \STATE Concept Drift Detected
	    \ENDIF
		\end{algorithmic}
		\label{alg:method}
	\end{algorithm}
\end{center}

In order to discover which concepts are shared, new, or missing we use the {\em Cauchy-Schwarz inequality} which states for two vectors \textbf{a} and \textbf{b} we have 
$\textbf{a}^T\textbf{b} \le ||\textbf{a}||_2 ||\textbf{b}||_2$.
Algorithm \ref{Algorithm: Find concepts} provides the general outline of technique used in finding concepts. It takes a column-normalized matrices $\textbf{A}_\textbf{old}$ and $\textbf{A}_\textbf{batch}$ of size $I\times R$ and $I \times batchRank$ respectively as input. We compute the dot product for all permutations of columns between two matrices, as shown below $$\textbf{A}^T_\textbf{old}(:,col_i) \cdot \textbf{A}_\textbf{batch}(:,col_j)$$
$col_i$ and $col_j$ are the respective columns.
If the computed dot product is higher than the threshold value, the two concepts match, and we consider them as shared/overlapping  between $\textbf{A}_\textbf{old}$ and $\textbf{A}_\textbf{batch}$. If the dot product between a column in $\textbf{A}_\textbf{batch}$ and with all the columns in $\textbf{A}_\textbf{old}$ has a value less than the threshold,  we consider it as a new concept. This solves problem \textbf{P2}. In the experimental evaluation, we demonstrate the behavior of \method with respect to that threshold.

\textbf{SeekAndDestroy:} 	This is our overall proposed algorithm, which detects concept drift between the two consecutive tensor batch decompositions, as illustrated in Algorithm \ref{Algorithm: Seek And Destroy} and updates the decomposition in a fashion robust to the drift. \method takes factor matrices($\textbf{A}_\textbf{old}$, $\textbf{B}_\textbf{old}$, $\textbf{C}_\textbf{old}$) of previous tensor batch (say at time $t_0$), running rank at $t_0$(\textbf{$\rr_{t_0}$}) and new tensor batch(\textbf{$\tensor{X}_{new}$}) (say at time $t_1$) as inputs. Subsequently, \method computes the tensor  rank for the batch (\textbf{batchRank}) for $\tensor{X}_{new}$ using AutoTen. 

Using the estimated rank \textbf{batchRank}, \method computes the CP decomposition of $\tensor{X}_{new}$, which returns factor matrices $\textbf{A},\textbf{B},\textbf{C}$. We normalize the columns of $A,B,C$ to unit $\ell_2$ norm and we store the normalized matrices into $\textbf{normMatA}$, $\textbf{normMatB}$, and $\textbf{normMatC}$, as shown by lines 3-4 of Algorithm \ref{Algorithm: Seek And Destroy}. Both $\textbf{A}_\textbf{old}$ and normalized matrix $\textbf{A}$ are passed to $findConceptOverlap$ function as described above. This returns the indexes of new concept and indexes of overlapping concepts from both matrices. Those indexes inform \method, while updating the factor matrices, where to append the overlapped concepts. If there are new concepts, we update $A$ and $B$ factor matrices simply by adding new columns from normalized factor matrices of $\tensor{X}_{new}$ as shown in lines 9-10 of Algorithm \ref{Algorithm: Seek And Destroy}. Furthermore,  we update the running rank by adding number of new concept discovered to the previous running rank. If there is only overlapping concepts and no new concepts, then $\mathbf{A}$ and $\mathbf{B}$ factor matrices does not change.

\textbf{Updating Factor Matrix C}:
In this paper, for simplicity of exposition, we are focusing on streaming data that are increasing only on one mode. However, our proposed method readily generalizes to cases where more than one modes grow over time. 

In order to update the ``evolving'' factor matrix ($\mathbf{C}$ in our case), we use a different technique from the one used to update $\mathbf{A}$ and $\mathbf{B}$. If there is a new concept discovered in $\textbf{normMatC}$ then 

\begin{equation}
\mathbf{C}_{updated}= \begin{bmatrix}
\mathbf{C}_{old}       & zeroCol \\
zerosM       & \mathbf{normMatC}(:,newConcept) \\
\end{bmatrix}
\end{equation}

where $zeroCol$ is of size $K_{old} \times len(newConcept)$, $zerosM$ is of size $K_{new} \times R$ and $\mathbf{C}_{updated}$ is of size $(K_{old}+K_{new}) \times \rr$.\\
If there are overlapping concepts, then we update $\mathbf{C}$ accordingly as shown below; in this case $\mathbf{C}_{updated}$ is again of size $(K_{old}+K_{new}) \times \rr$.\\

\begin{equation}
\mathbf{C}_{updated}= \begin{bmatrix}
\mathbf{C}_{old}(:,overlapConceptOld)\\
\mathbf{normMatC}(:,conceptOverlap) \\
\end{bmatrix}
\end{equation}

If there are missing concepts we append an all-zeros matrix (column vector) to those indexes.

\textbf{The Scaling Factor $\rho$:} When we reconstruct the tensor from updated factor (normalized) matrices, we need a way to re-scale the columns of those factor matrices. In our approach we compute element wise product on normalized columns of factor matrices ($\mathbf{A}$, $\mathbf{B}$, $\mathbf{C}$) of $\tensor{X}_{new}$ as shown in line 5 of Algorithm \ref{Algorithm: Seek And Destroy}. We use the same technique as the one used in updating C matrix, in order to match the values between two consecutive intervals, and we add this value to previously computed values. If it is a missing concept, we simply add zero to it.  While reconstructing the tensor we take the average of vector over the number of batches received and we re-scale the components as follows
\[
	\tensor{X}_r  = \displaystyle{\sum_{r=1}^{\rr} \rho_r \mathbf{A}_\text{upd.}(:,r) \circ \mathbf{B}_\text{upd.}(:,r) \circ \mathbf{C}_\text{upd.}(:,r)}.
\]

\begin{center}
	\begin{algorithm} [!ht]
		\caption{Find Concept Overlap}
		\label{Algorithm: Find concepts}
		\begin{algorithmic} [1]
			\REQUIRE Factor matrices $\mathbf{A}_{old},\mathbf{normMatA }$ of size $I \times R$, $I \times batchRank$ respectively.
			\ENSURE \newcon, \conOverlap, overlapConceptOld
			
			\STATE $THRESHOLD \leftarrow 0.6$
			\IF{$R == batchRank$}
			\STATE Generate all the permutations for [1:R]
			\STATE \ForEach{permutation}{Compute dot product of $\mathbf{A}_{old}\ and\ \mathbf{normMatA(:,permutation)}$}
			
			\ELSIF{$R > batchRank$}
			\STATE Generate all the permutations(1:R, batchRank)
			\STATE \ForEach{permutation}{Compute dot product of $\mathbf{A}_{old}(:,permutation)\ and\ \mathbf{normMatA}$}

			\ELSIF{$R < batchRank$}
			\STATE Generate all the permutations (1:batchRank, R)
			\STATE \ForEach{permutation}{Compute dot product of $\mathbf{A}_{old}\ and\ \mathbf{normMatA(:,permutation)}$}
			\ENDIF
			\STATE Select the best permutation based on the maximum sum.
			\STATE If dot product value of a column is less than threshold its a \newconcept 
			\STATE If dot product value of a column is more than threshold then its a \conceptoverlap. 
			\STATE Return column index's of \newconcept and \conceptoverlap  for both matrices
		\end{algorithmic}
		\label{alg:method}
	\end{algorithm}
\end{center}

\section{Experimental Evaluation}
\label{sec:experiments}
We evaluate our algorithm on the following criteria:\\
{\bf Q1: Approximation Quality}: We compare \method's reconstruction accuracy against state-of-the-art streaming baselines, in data that we generate synthetically so that we observe different instances of concept drift. In cases where \method outperforms the baselines, we argue that this is due to the detection and alleviation of concept drift. \\
{\bf Q2: Concept Drift Detection Accuracy}: We evaluate how effectively \method is able to detect concept drift in synthetic cases, where we control the drift patterns.\\
{\bf Q3: Sensitivity Analysis}: As shown in Section \ref{sec:method}, \method expects the matching threshold as a user input. Furthermore, its performance may depend on the selection of the batch size. Here, we experimentally evaluate \method's sensitivity along those axes. \\
{\bf Q4: Effectiveness on Real Data}: In addition to measuring \method's performance in real data, we also evaluate its ability to identify useful and interpretable latent concepts in real data, which elude other streaming baselines. \\
\subsection{Experimental Setup}
We implemented our algorithm in Matlab using tensor toolbox library \cite{TTB_Software} and we evaluate our algorithm on both synthetic and real data.We use ~\cite{papalexakis2016automatic} method available in literature to find rank of incoming batch.

In order to have full control of the drift phenomena, we generate synthetic tensors with different ranks for every tensor batch, we control the batch rank of the tensor with factor matrix \textbf{C}.  Table \ref{table:tsyndataset} shows the specification of the datasets created. For instance dataset \textbf{SDS2} has an initial tensor batch whose tensor rank is $2$ and last tensor batch whose tensor rank is $10$(full rank). The batches in between the initial and final tensor batch can have any rank between initial and final rank(in this case 2-10). The reason we assign the final batch rank as the full rank is to make sure the tensor created is not rank deficient. We make the synthetic tensor generator available as part of our code release.

\begin{table*}[h!]
	\ssmall
	\begin{center}
		\begin{tabular}{ |c||c|c|c|c|c|c|}
			\hline
			DataSet & Dimension & Initial Rank &  Full Rank  &Batch Size&Matching Threshold  \\
			\hline
			\hline
			SDS1&\multirow{ 2}{*}{100 x 100 x 100} &\multirow{ 2}{*}{2}&5 &\multirow{ 2}{*}{10}&\multirow{ 2}{*}{0.6} \\
		     SDS2& & &10 & &  \\
		      \hline
		      \hline
		    SDS3&\multirow{ 2}{*}{300 x 300 x 300} &\multirow{ 2}{*}{2}&5 &\multirow{ 2}{*}{50}&\multirow{ 2}{*}{0.6} \\
		    SDS4&   & &10 & &  \\
		      \hline
		      \hline
		   SDS5& \multirow{ 2}{*}{500 x 500 x 500} &\multirow{ 2}{*}{2}&5 &\multirow{ 2}{*}{100}&\multirow{ 2}{*}{0.6} \\
		   SDS6&  & &10 & &  \\
		     \hline
		\end{tabular}
		\bigskip
		\caption{Table of Datasets analyzed}
		\label{table:tsyndataset}
	\end{center}
\end{table*}

In order for us to obtain robust estimates of performance, we require all experiments to either 1) run for 1000 iterations, or 2) the standard deviation converges to a second significant digit (whichever occurs first). For all reported results, we use the median and the standard deviation.

\subsection{Evaluation Metrics}

We evaluate \method and the baselines methods using {\em relative error}. Relative Error provides the measure of effectiveness of the computed tensor with respect to the original tensor and is defined as follows (lower is better):
\begin{equation}
Relative Error= \Big(\frac{||\tensor{X}_{original}-\tensor{X}_{computed}||_F}{||\tensor{X}_{original}||_F}\Big)
\end{equation}

\subsection{Baselines for Comparison}
To evaluate our method, we compare \method with two state-of-the-art streaming baselines: OnlineCP \cite{zhou2016accelerating} and SamBaTen \cite{gujral2017sambaten}. Both baselines assume that the rank remains fixed throughout the entire stream. When we evaluate the approximation accuracy of the baselines, we run two different versions of each method, with different input ranks: 1) {\em Initial Rank}, which is the rank of the initial batch, same as the one that \method uses, and 2) {\em Full Rank}, which is the ``oracle'' rank of the full tensor, if we assume we could compute that in the beginning of the stream. Clearly, {\em Full Rank} offers a great advantage to the baselines since it provides information from the future.

\subsection{Q1: Approximation Quality}
The first dimension that we evaluate is the approximation quality. More specifically, we evaluate whether \method is able to achieve good approximation of the original tensor (in the form of low error) in case where concept drift is occurring in the stream. Table \ref{table:resulttable} contains the general results of \method's accuracy, as compared to the baselines. We observe that \method outperforms the two baselines, in the pragmatic scenario where they are given the same starting rank as \method (Initial Rank). In the non-realistic, ``oracle'' case, OnlineCP performs better than SamBaTen and \method, however this case is a very advantageous lower bound on the error for OnlineCP.

\begin{table*}[h!]
	\ssmall
	\begin{center}
		\begin{tabular}{ |p{1cm}|p{1.7cm}|p{1.7cm}|p{1.5cm}|p{1.5cm}|p{2cm}|}
			\hline
			\centering{DataSet}  & \centering{OnlineCP (Initial Rank)} & \centering{OnlineCP (Full Rank)} &  \centering{SamBaTen (Initial Rank)} & \centering{SamBaTen (Full Rank)} & \method \\
			\hline
			SDS1&  0.2782$\pm$0.0221 & \textcolor{red}{0.197$\pm$0.086} & \textbf{0.261$\pm$0.048}& 0.317$\pm$0.058 & 0.283$\pm$0.075\\
			SDS2& 0.2537$\pm$0.0125&\textcolor{red}{0.168$\pm$0.507}& \textbf{0.244$\pm$0.028}&0.480$\pm$0.051 & 0.253$\pm$0.0412\\
			SDS3&  0.2731$\pm$0.0207&\textcolor{red}{0.205$\pm$0.164}& 0.385$\pm$0.021& 0.445$\pm$0.164 & \textbf{0.266$\pm$0.081}\\
			SDS4&  0.245$\pm$0.013&\textcolor{red}{0.171$\pm$0.537}& 0.299$\pm$0.045& 0.402$\pm$0.049 & \textbf{0.221$\pm$0.0423}\\
			SDS5&  0.2719$\pm$0.0198&\textcolor{red}{0.206$\pm$0.022}& 0.559$\pm$0.046& 0.519$\pm$0.0219 & \textbf{0.256$\pm$0.105}\\
			SDS6&  0.238$\pm$0.013&\textcolor{red}{0.171$\pm$0.374}& 0.510$\pm$0.036& 0.547$\pm$0.0276` & \textbf{0.208$\pm$0.0433}\\
			\hline	
	
		\end{tabular}
		\bigskip
		\caption{Approximation error for \method and the baselines. \method outperforms the baselines in the realistic case where all methods start with the same rank}
		\label{table:resulttable}
	\end{center}
	\vspace{-0.4in}
\end{table*}

Through extensive experimentation we made the following interesting observation: in the cases where most of the concepts in the stream appear in the beginning of the stream (e.g., in batches 2 and 3), \method was able to further outperform the baselines. This is due to the fact that, if \method has already ``seen'' most of the possible concepts early-on in the stream, it is more likely to correctly match concepts in later batches of the stream, since there already exists an almost-complete set of concepts to compare against. Indicatively,in this case \method achieved $0.1176 \pm 0.0305$ where as OnlineCP achieved $0.1617 \pm 0.0702$.

\hide{\begin{figure}[!ht]
\begin{center}
	\includegraphics[clip,width = 0.45\textwidth]{FIG/error_special.eps}
	\caption{In cases where most of the existing components in the stream are discovered early-on, \method outperforms OnlineCP with a bigger margin.}
	\label{fig:special}
\end{center}
\end{figure}}

\subsection{Q2: Concept Drift Detection Accuracy}
\begin{figure}[!ht]
	\vspace{-0.2in}
	\begin{center}
		\subfigure[Increasing rank]{\includegraphics[clip,width = 0.45\textwidth]{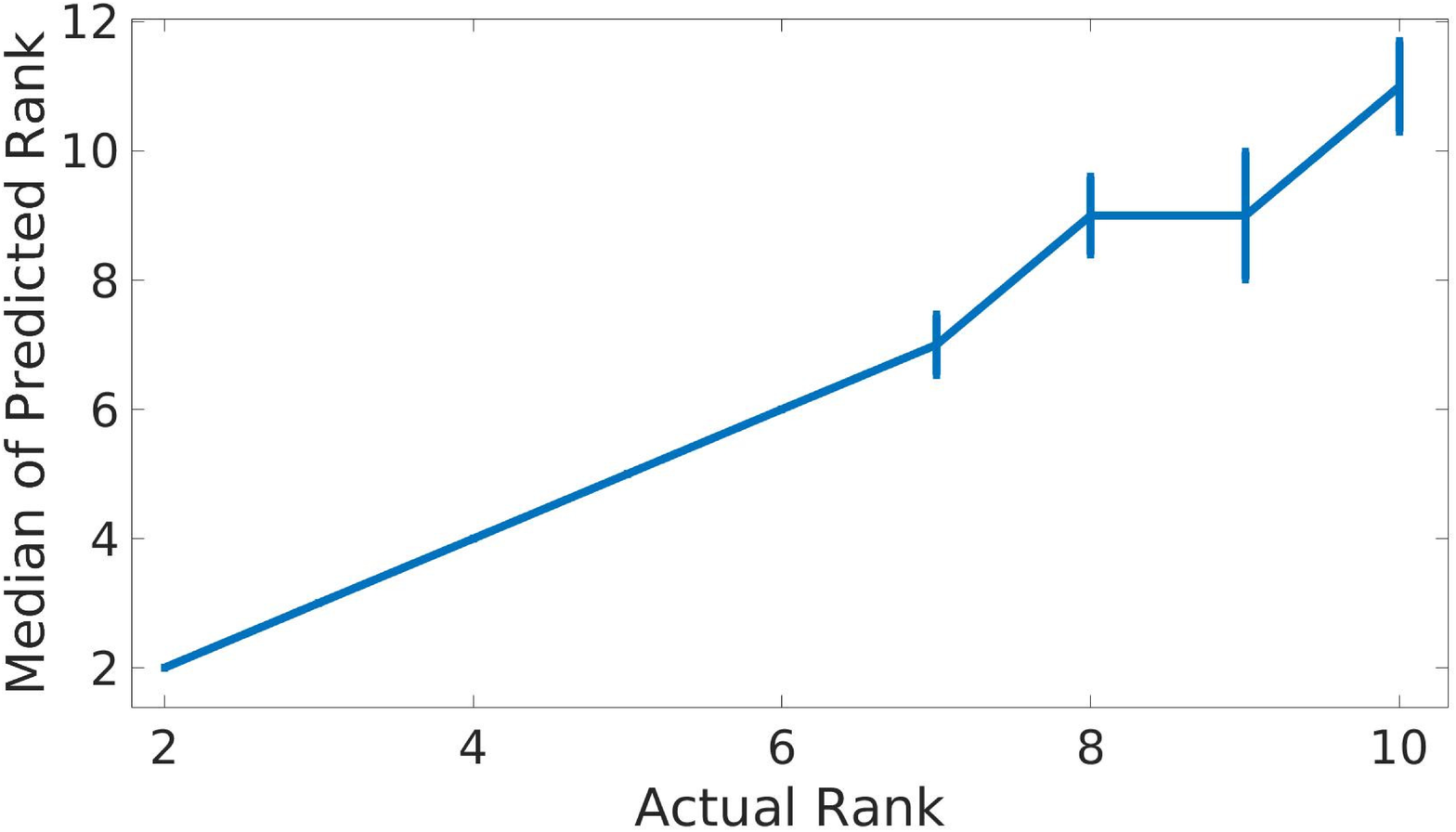}}
		\subfigure[Decreasing rank]{\includegraphics[clip,width = 0.45\textwidth]{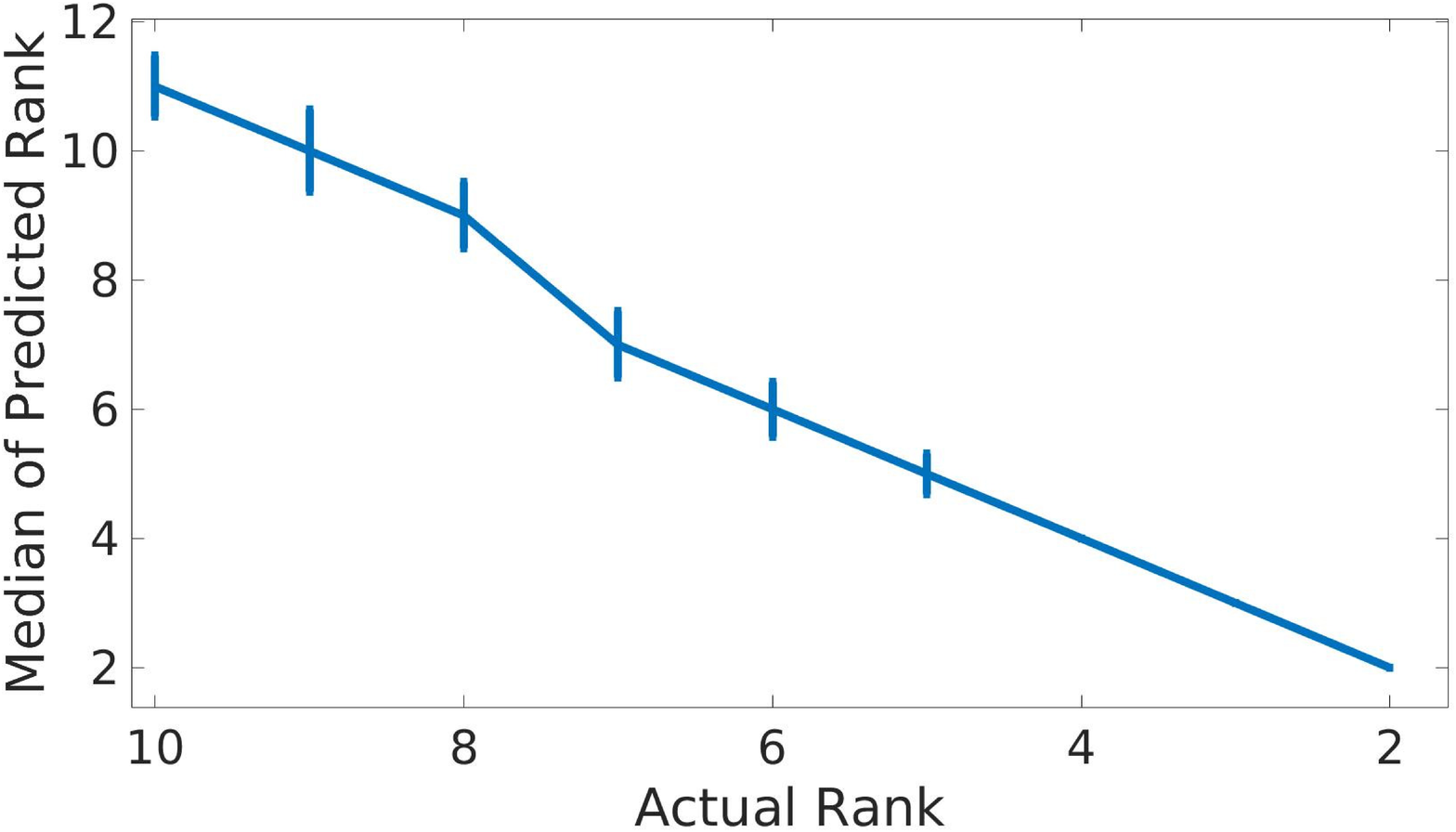}}
		\caption{\method is able to successfully detect concept drift, which is manifested as changes in the rank throughout the stream}
		\label{fig:detection_accuracy}
	\end{center}
	\vspace{-0.1in}
\end{figure}

The second dimension along which we evaluate \method is its ability to successfully detect concept drift. Figure \ref{fig:detection_accuracy} shows the rank discovered by \method at every point of the stream, plotted against the actual rank. We observe that \method is able to successfully identify changes in rank, which, as we have already argued, signify concept drift. Furthermore, Table \ref{tbl:mtachthreshold}(b) shows three example runs that demonstrate the concept drift detection accuracy.

\subsection{Q3: Sensitivity Analysis}
The results we have presented so far for \method have used a matching threshold of 0.6. The threshold was chosen because it is intuitively larger than a 50\% match, which is a reasonable matching threshold. In this experiment, we investigate the sensitivity of \method to the matching threshold parameter. Table \ref{tbl:mtachthreshold}(a) shows exemplary approximation errors for thresholds of 0.4, 0.6, and 0.8. We observe that 1) the choice of threshold is fairly robust for values around 50\%, and 2) the higher the threshold, the better the approximation, with threshold of 0.8 achieving the best performance.

\begin{table}
	\ssmall
	\begin{center}
\begin{tabular}{cc}
 	\begin{minipage}{.4\linewidth}
		\begin{tabular}{ |c||c|c|c|  }
			\hline
			Threshold & SDS2 &  SDS4     \\
			\hline
			\hline
			 0.4  &0.253$\pm$0.041 & 0.221 $\pm$ 0.042 \\
	 
			\hline
			\hline
		     0.6 &0.253$\pm$0.041 &0.221 $\pm$ 0.042  \\
		
			\hline
			\hline
			 0.8 &0.101 $\pm$0.040&0.033 $\pm$ 0.011   \\
		
			\hline
		\end{tabular}
 	\end{minipage} &
 	\hspace{-0.1in}
 	\begin{minipage}{.6\linewidth}
 	\begin{tabular}{|c|c|c|c|c|c|}
 		\hline
 		\multirow{1}{*} Running & Actual &  Predicted  & \multicolumn{2}{|c|}{\textbf{Approx. Error}}   \\
 		Rank&Rank&Rank&Actual&Predicted\\
 		&&&Rank & Rank\\
 	    \hline
 		\hline
 		6&[2,4,3,4,3,3,5,3,3,5]&[2,4,3,4,3,3,5,3,3,6]&0.185&0.194 \\
 		6&[2,4,3,4,3,3,5,3,3,5]&[2,4,3,4,3,3,5,3,3,6]&0.185&0.197\\
 		7&[2,4,3,4,3,3,5,3,3,5]&[2,4,3,5,3,3,6,3,3,6]&0.185&0.278 \\
 	 	\hline
 	\end{tabular}
 \end{minipage} 

\end{tabular}
\end{center}
\caption{(a)Experimental results for error of approximation of incoming batch with different matching threshold values. Dataset SDS2 and SDS4 are of dimension $\mathbb{R}^{100 \times 100 \times 100}$ and $\mathbb{R}^{300 \times 300 \times 300}$ , respectively. We see that the threshold is fairly robust around 0.5, and a threshold of 0.8 achieves the highest accuracy (b) Experimental results on SDS1 for error of approximation of incoming slices with known and predicted rank}
\label{tbl:mtachthreshold}
\end{table}

\begin{table}[ht]
	\begin{center}
	\begin{tabular}{ |c|c|c|c|c|c| }
		\hline
		\multirow{ 1}{*}Running &Predicted &  Batch  & \multicolumn{3}{|c|}{\textbf{Approximation Error}}   \\
		Rank&Full Rank&Size&\method& SambaTen& OnlineCP  \\
		\hline
		\hline
		7$\pm$0.88&4$\pm$0.57&22&\textbf{0.68 $\pm$ 0.002} &0.759$\pm$ 0.059&0.941$\pm$ 0.001\\
		
		\hline
	\end{tabular}
	\end{center}
\caption{Evaluation on Real dataset}
\label{tbl:enronEvaluation}
\end{table}

\subsection{Q4: Effectiveness on Real Data}
To evaluate effectiveness of our method on real data, we use the Enron time-evolving communication graph dataset \cite{bader2006analysis}. Our hypothesis is that in such complex real data, there should exists concept drift in streaming tensor decomposition. In order to validate that hypothesis, we compare the approximation error incurred by \method against the one incurred by the baselines, shown in Table \ref{tbl:enronEvaluation}. We observe that the approximation error of \method is lower than the two baselines. Since the main difference between \method and the baselines is that \method takes concept drift into consideration, and strives to alleviate its effects, this result 1) provides further evidence that there exists concept drift in the Enron data, and 2) demonstrates \method's effectiveness on real data.
\begin{figure}[!ht]
	\includegraphics[width=1\textwidth]{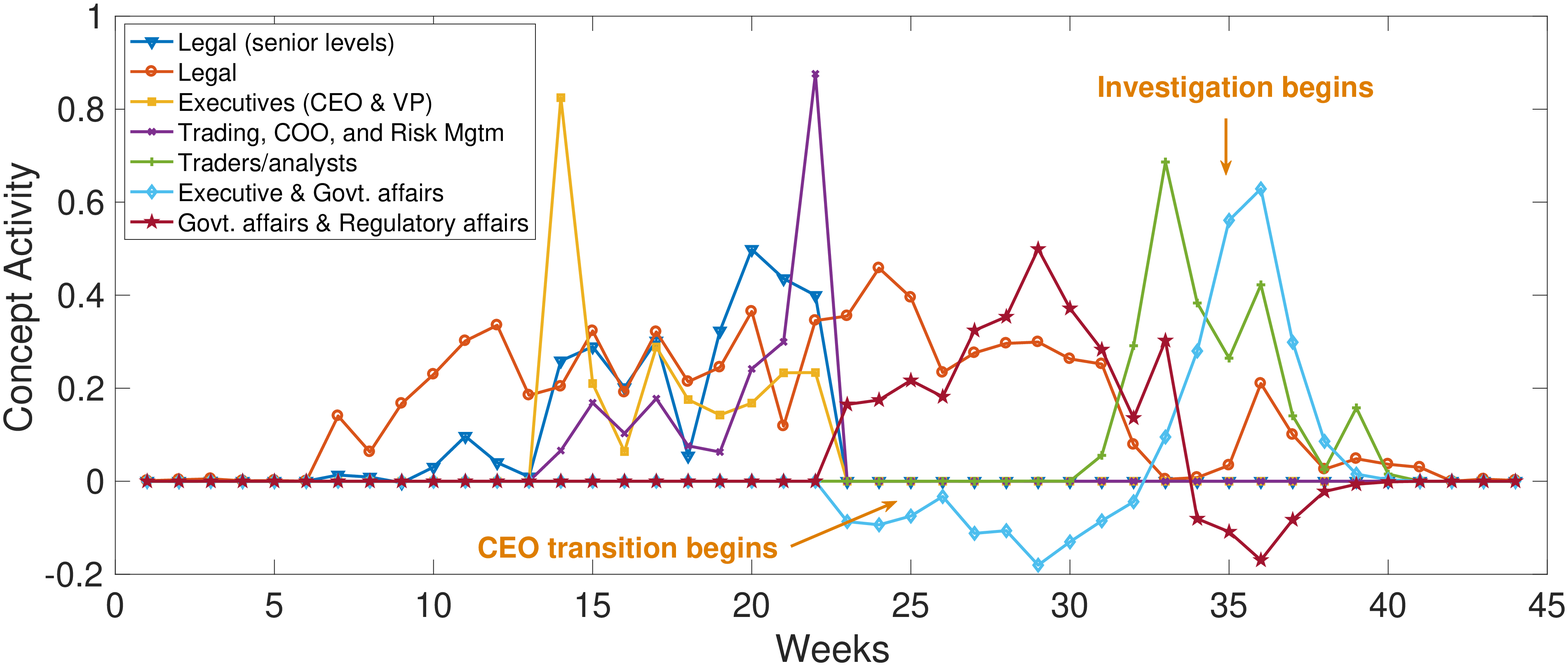}
	\caption{Timeline of concepts discovered in Enron}
\end{figure}	

The final rank for Enron as computed by \method was 7, indicating the existence of 7 time-evolving communities in the dataset. This number of communities is higher than what previous tensor-based analysis has uncovered \cite{bader2006analysis,papalexakis2012parcube}. However, analyzing the (static) graph using a highly-cited non-tensor based method \cite{blondel2008fast}, we were able to detect 7 communities, therefore \method may be discovering subtle communities that have eluded previous tensor analysis. In order to verify that, we delved deeper into the communities and we plot their temporal evolution (taken from matrix $\mathbf{C}$) along with their annotations (when inspecting the top-5 senders and receivers within each community). Indeed, a subset of the communities discovered matches with the ones already known in the literature \cite{bader2006analysis,papalexakis2012parcube}. Additionally, \method was able to discover community \#3, which refers to a group of executives, including the CEO. This community appears to be active up until the point that the CEO transition begins, after which point it dies out. This behavior is indicative of concept drift, and \method was able to successfully discover and extract it.

\hide{
\begin{SCfigure}
	\vspace{-0.1in}
	\includegraphics[clip,trim=0cm 3.8cm 0cm 2.1cm,width = 0.35\textwidth]{FIG/enron_communities.png}
	\caption{Communities discovered in enron dataset using gephi tool\cite{ICWSM09154}}
	\label{fig:sen_shared}
\end{SCfigure}
}
\section{Related Work}
\label{sec:related}

\textbf{Tensor decomposition:} Tensor decomposition techniques are widely used for static data. With the explosion of big data, data grows at a rapid speed and an extensive study required on the online tensor decomposition problem.  Sidiropoulos \cite{nion2009adaptive} introduced two well-known PARAFAC based methods namely RLST (recursive least square) and SDT (simultaneous diagonalization tracking) to address the online 3-mode tensor decomposition. Zhou et al. \cite{zhou2016accelerating} proposed OnlineCP for accelerating online factorization that can track the decompositions when new updates arrived for N-mode tensors.  Gujral et al. \cite{gujral2017sambaten} proposed Sampling-based Batch Incremental Tensor Decomposition algorithm which updates online computation of CP/PARAFAC and performs all computations in the reduced summary space. However, no prior work addresses concept drift.

\textbf{Concept Drift:} The survey paper \cite{Webb2016} provides the qualitative definitions of characterizing the drifts on data stream models. To the best of our knowledge, however, this is the first work to discuss concept drift in tensor decomposition.
\section{Conclusions}
\label{sec:conclusions}
In this paper we introduce the notion of ``concept drift'' in streaming tensors. and provide \method, an algorithm which detects and alleviates concept drift it without making any assumption on the rank of the tensor. \method outperforms other state-of-the-art methods when the rank is unknown and is effective in detecting concept drift. Finally, we apply \method on a real time-evolving dataset, discovering novel drifting concepts.

\section*{Acknowledgements}
{
Research was supported by the Department of the Navy, Naval Engineering Education Consortium under award no. N00174-17-1-0005, the National Science Foundation EAGER Grant no. 1746031, and by an Adobe Data Science Research Faculty Award. Any opinions, findings, and conclusions or recommendations expressed in this material are those of the author(s) and do not necessarily reflect the views of the funding parties.
}

\balance
\bibliographystyle{splncs04}

\end{document}